\pdfoutput=1

\documentclass[11pt]{article}

\usepackage{acl}

\usepackage{times}
\usepackage{latexsym,booktabs}
\usepackage{graphicx}
\usepackage{xcolor}

\usepackage[T1]{fontenc}

\usepackage[utf8]{inputenc}

\usepackage{microtype}
\usepackage{xspace}

\newcommand{\bias}{{\sc Square One Bias}\xspace}
\newcommand{\biasupper}{Square One Bias\xspace}
\newcommand{\point}{{\sc Square One}\xspace}
\newcommand{\pointupper}{Square One\xspace}
\newcommand{\pointlower}{square one\xspace}

\newcommand{\rparagraph}[1]{\vspace{1.6mm}\noindent\textbf{#1.}}

\title{Square One Bias in NLP:\\Towards a Multi-Dimensional Exploration of the Research Manifold}

\author{Sebastian Ruder\thanks{$\;$ The authors contributed equally to this work.} \\
  Google Research \\
  \texttt{ruder@google.com} \\\And
  Ivan Vulić\footnotemark[1] \\
  University of Cambridge \\
  \texttt{iv250@cam.ac.uk} \\\And
  Anders Søgaard\footnotemark[1] \\
  University of Copenhagen \\
  \texttt{soegaard@di.ku.dk}}

\begin{document}
\maketitle
\begin{abstract}
The prototypical NLP experiment trains a {standard architecture} on labeled {\bf English} data and optimizes for {\bf accuracy}, without accounting for other dimensions such as {\bf fairness}, {\bf interpretability}, or computational {\bf efficiency}. We show through a manual classification of recent NLP research papers that this is indeed the case and refer to it as the {\em \pointlower} experimental setup. We observe that NLP research often goes beyond the \pointlower setup, e.g, focusing not only on accuracy, but also on fairness or interpretability, but typically {\em only} along a single dimension. Most work targeting multilinguality, for example, considers only accuracy; most work on fairness or interpretability considers only English; and so on.
We show this through manual classification of recent NLP research papers and ACL Test-of-Time award recipients.
Such one-dimensionality of most research means we are only exploring a fraction of the NLP research search space. We provide historical and recent examples of how the \pointlower bias has led researchers to draw false conclusions or make unwise choices, point to promising yet unexplored directions on the research manifold, and make practical recommendations to enable more multi-dimensional research. We open-source the results of our annotations to enable further analysis.\footnote{\url{https://github.com/google-research/url-nlp}}
\end{abstract}

\section{Introduction}

\begin{figure}[ht!]
	\centering
	\includegraphics[width=1.05\linewidth]{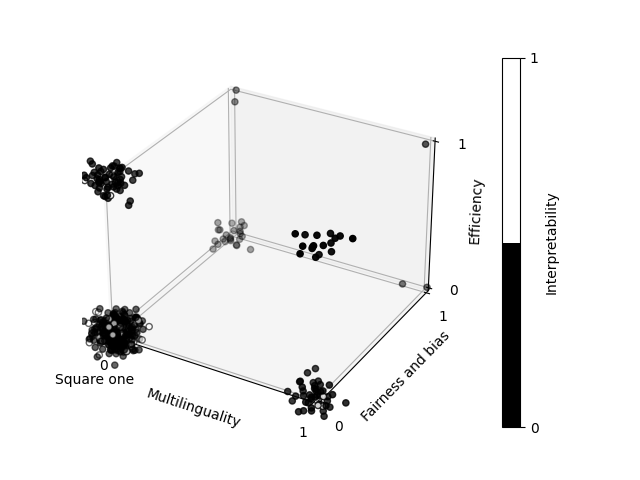}
	\caption{Visualization of contributions of ACL 2021 oral papers along 4 dimensions: multilinguality, fairness and bias, efficiency, and interpretability (indicated by color). Most work is clustered around the \point or along a single dimension.}
	\label{fig:acl_papers_2021_4d}
\end{figure}

Our categorization of objects, say screwdrivers or NLP experiments, is heavily biased by early prototypes \cite{sherman85,DASSMAAL1990349}. If the first 10 screwdrivers we see are red and for hexagon socket screws, this will bias what features we learn to associate with screwdrivers. Likewise, if the first 10 NLP experiments we see or conduct are in sentiment analysis, this will likely also bias how we think of NLP experiments in the future.  

In this position paper, we postulate that we can meaningfully talk about {\em the} {prototypical}~NLP experiment, and that {\em the existence of such an experimental prototype steers and biases the research dynamics in our community}. We will refer to this prototype as NLP's \point---and to the bias that follows from it, as the \bias. We argue this bias manifests in a particular way: Since research is a creative endeavor, and researchers aim to push the research horizon, {\em most research papers in NLP go beyond this prototype, but only along a single dimension at a time}. Such dimensions might include multilinguality, efficiency, fairness, and interpretability, among others. The effect of the \bias is to baseline novel research contributions, rewarding work that differs from the prototype in a concise, one-dimensional way.

We present several examples of this effect in practice. For instance, analyzing the contributions of ACL 2021 papers along 4 dimensions, we observe that most work is either clustered around the \point or makes a contribution along a single dimension (see Figure \ref{fig:acl_papers_2021_4d}).
Multilingual work typically disregards efficiency, fairness, and interpretability. Work on efficient NLP typically only performs evaluations on English datasets, and disregards fairness and interpretability. Fairness and interpretability work is also mostly limited to English, and tends to disregard efficiency concerns.

We argue that the \bias has several negative effects, most of which amount to the study of one of the above dimensions being biased by ignoring the others. Specifically, by focusing only on exploring the edges of the manifold, we are not able to identify the non-linear interactions between different research dimensions. We highlight several examples of such interactions in Section~\ref{sec:examples}. 
Overall, we encourage a focus on combining multiple dimensions on the research manifold in future NLP research, and delve deeper into studying their (linear and non-linear) interactions.

\rparagraph{Contributions} We first establish that we can meaningfully talk about the prototypical NLP experiment, through a series of annotation experiments and surveys. This prototype amounts to applying a standard architecture to an English dataset and optimizing for accuracy or F1. We discuss the impact of this prototype on our research community, and the bias it introduces. We then discuss the negative effects of this bias. We also list work that has taken steps to overcome the bias. Finally, we highlight blind spots and unexplored research directions and make practical recommendations, aiming to inspire the community towards conducting more `multi-dimensional' research (see Figure~\ref{fig:acl_papers_2021_4d}).

\section{Finding the \pointupper}

In order to determine the existence and nature of a \point, we assess contemporary research in NLP along a number of different dimensions.

\rparagraph{Dimensions} We identify potential themes in NLP research by reviewing the Call for Papers, publication statistics by area, and paper titles of recent NLP conferences. We focus on \emph{general} dimensions that are not tied to a particular task and are applicable to any NLP application.\footnote{For instance, we do not consider multimodality, as a task or model is inherently multimodal or not.} We furthermore focus on dimensions that are represented in a reasonable fraction of NLP papers (at least 5\% of ACL 2021 oral papers).\footnote{Privacy, interactivity, and other emerging research areas are excluded based on this criterion.} Our final selection focuses on 4 dimensions along which papers may make research contributions: multilinguality, fairness and bias, efficiency, and interpretability. Compared to prior work that annotates the values of ML research papers \cite{Birhane2021}, we are not concerned with a paper's motivation but whether its \emph{practical contributions} constitute a meaningful departure from the \point. For each paper, we annotate whether it makes a contribution along each dimension as well as the languages and metrics it employs for evaluation. We provide the detailed annotation guidelines in Appendix \ref{app:annotation_guidelines}.


\begin{table*}[]
\centering
\resizebox{\textwidth}{!}{%
\begin{tabular}{l ccccccccc}
\toprule
Area & \# papers & & English & Accuracy / F1 & Multilinguality & Fairness and bias & Efficiency & Interpretability & \textgreater{}1 dimension \\ \midrule
ACL 2021 oral papers & 461 & & 69.4\% & 38.8\% & 13.9\% & 6.3\% & 17.8\% & 11.7\% & 6.1\% \\ \midrule
MT and Multilinguality & 58 &  & 0.0\% & 15.5\% & 56.9\% & 5.2\% & 19.0\% & 6.9\% & 13.8\% \\
Interpretability and Analysis & 18 &  & 88.9\% & 27.8\% & 5.6\% & 0.0\% & 5.6\% & 66.7\% & 5.6\% \\
Ethics in NLP & 6 &  & 83.3\% & 0.0\% & 0.0\% & 100.0\% & 0.0\% & 0.0\% & 0.0\% \\ \midrule
Dialog and Interactive Systems & 42 &  & 90.5\% & 21.4\% & 0.0\% & 9.5\% & 23.8\% & 2.4\% & 2.4\% \\
Machine Learning for NLP & 42 &  & 66.7\% & 40.5\% & 19.0\% & 4.8\% & 50.0\% & 4.8\% & 9.5\% \\
Information Extraction & 36 &  & 80.6\% & 91.7\% & 8.3\% & 0.0\% & 25.0\% & 5.6\% & 8.3\% \\
Resources and Evaluation & 35 &  & 77.1\% & 42.9\% & 5.7\% & 8.6\% & 5.7\% & 14.3\% & 5.7\% \\
NLP Applications & 30 &  & 73.3\% & 43.3\% & 0.0\% & 10.0\% & 20.0\% & 10.0\% & 0.0\% \\
\bottomrule
\end{tabular}%
}
\caption{The number of ACL 2021 oral papers (top row) and of papers in each area (bottom rows) as well as the fractions that only evaluate on English, only use accuracy / F1, make contributions along one of four dimensions, and make contributions along more than a single dimension (from left to right).}
\label{tab:acl2021}
\end{table*}

\rparagraph{ACL 2021 Oral Papers} We annotate the 461 papers that were presented orally at ACL 2021, a representative cross-section of the 779 papers accepted to the main conference. The general statistics from our classification of ACL 2021 papers are presented in Table~\ref{tab:acl2021}. In addition, we highlight the statistics for the conference {\em areas} (tracks) corresponding to 3 of the 4 dimensions\footnote{Unlike EACL 2021, NAACL-HLT 2021 and EMNLP 2021, ACL 2021 had no area associated with efficiency. To compensate for this, we annotated the 20 oral papers of the ``Efficient Models in NLP'' track at EMNLP 2021 (see Appendix \ref{app:efficiency_emnlp2021}).}, as well as for the top 5 areas with the most papers. We show statistics for the remaining areas in Appendix \ref{app:acl2021_remaining}. We additionally visualize their distribution in Figure \ref{fig:acl_papers_2021_4d}. Overall, almost 70\% of papers evaluate only on English, clearly highlighting a lack of language diversity in NLP \cite{bender2011achieving,Joshi2020}. Almost 40\% of papers only evaluate using accuracy and/or F1, foregoing metrics that may shed light on other aspects of model behavior. 56.6\% of papers do not study any of the four major dimensions that we investigated. We refer to this standard experimental setup---evaluating only on \textbf{English} and optimizing for \textbf{accuracy} or another performance metric without considering other dimensions---as the \point.

Regarding work that moves from the \point, most papers make a contribution in terms of efficiency, followed by multilinguality. However, most papers that evaluate on multiple languages are part of the corresponding MT and Multilinguality track. Despite being an area receiving increasing attention \cite{Blodgett2020}, only 6.3\% of papers evaluate the bias or fairness of a method. Overall, \textit{only 6.1\% of papers} make a contribution along two or more of these dimensions. Among these, joint contributions on both multilinguality and efficiency are the most common (see Figure \ref{fig:acl_papers_2021_4d}). In fact, 22 of the 26 two-or-more-dimensional papers focus on efficiency, and 17 of these on the combination of multilinguality and efficiency. This means less than 1\% of the ACL 2021 papers consider combinations of (two or more of) multilinguality, fairness and interpretability. We find this surprising, given these topics are considered among the most popular topics in the field. 

Some areas have particularly concerning statistics. A large majority of research work in dialog (90.5\%), summarization (91.7\%), sentiment analysis (100\%), and language grounding (100\%) is done only on English; however, ways of expressing sentiment \cite{volkova-etal-2013-exploring,Yang:2017tacl,Vilares:2018sentic} and visually grounded reasoning \cite{Liu2021marvl,yin-etal-2021-broaden} do vary across languages and cultures. Systems in the top tracks tend to evaluate efficiency, but in general do not consider fairness or interpretability of the proposed methods. Even the creation of new resources and evaluation sets (cf., Resource and Evaluation in Table~\ref{tab:acl2021}) seems to be directed towards rewarding and enabling \point experiments; favoring English (77.1\%), and with modest efforts on other dimensions. Notably, we only identified a single paper that considers three dimensions \cite{renduchintala-etal-2021-gender}. This paper considers gender bias (Fairness) in relation to speed-quality (Efficiency) trade-offs in multilingual machine translation (Multilinguality). Finally, we observe that best-paper award winning papers are {\em not} more likely to consider more than one of the four dimensions. Only 1 in 8 papers did; the best paper \cite{xu-etal-2021-vocabulary}, like most two-dimensional ACL 2021 papers, considered multilinguality and efficiency.


\begin{table}[!t]
\footnotesize  
    \centering
    \def\arraystretch{0.93}
    \begin{tabular}{llll}
    \toprule
    {\bf Year}&{\bf Paper}&{\bf Language}&{\bf Metric}\\
    \midrule 
        1995&\citet{grosz-etal-1995-centering}&English&n/a\\
        1995&\citet{yarowsky-1995-unsupervised}&English&acc.\\
         1996&\citet{berger-etal-1996-maximum}&English&acc.\\  1996&\citet{carletta-1996-assessing}&n/a&n/a\\
         \midrule 
         2010&\citet{baroni-lenci-2010-distributional}&English&acc.\\
         2010&\citet{turian-etal-2010-word}&English&F$_1$\\
         2011&\citet{taboada-etal-2011-lexicon}&English&acc. \\
         2011&\citet{ott-etal-2011-finding}&English&acc./F$_1$\\
         \bottomrule 
    \end{tabular}
    \caption{Test-of-Time Award 2021-22 papers}
    \label{tab:testoftime}
\end{table}
\rparagraph{Test-of-Time Award Recipients} Current papers provide us with a snapshot of {\em actual} current research practices, but the one-dimensionality of the best paper award winning papers at ACL 2021 suggest the \bias~also biases what we {\em value} in research, i.e., our perception of {\em ideal}~research practices. This can also be seen in the papers that have received the ACL Test-of-Time Award in the last two years (Table~\ref{tab:testoftime}). Seven in eight papers included empirical evaluations performed exclusively on English data. Six papers were exclusively concerned with optimizing for accuracy or $F_1$. 

\rparagraph{Blackbox NLP Papers} Finally, we check if more multi-dimensional papers were presented at a workshop devoted to one of the above dimensions. The rationale is that if everyone at a workshop already explores one of these dimensions, including another may be a way to have an edge over other submissions. Unfortunately, this does not seem to be the case. We manually annotated the first 10 papers in the Blackbox NLP 2021 program\footnote{\url{https://blackboxnlp.github.io/}} that were available as pre-prints at the time of submission. Of the 10 papers, only one included more than one dimension \cite{abdullah2021familiar}. This number aligns well with the overall statistics of ACL 2021 (6.1\%). All the other Blackbox NLP papers only considered interpretability for English. 

\section{\biasupper: Examples} \label{sec:examples}

In the following, we highlight both historical and recent examples touching on different aspects of research in NLP that illustrate how the gravitational attraction of the \point has led researchers to draw false conclusions, unconsciously steer standard research practices, or make unwise choices. 

\rparagraph{Architectural Biases} 
One pervasive bias in our models regards \textbf{morphology}. Many of our models were not designed with morphology in mind, arguably because of the poor/limited morphology of English. Traditional n-gram language models, for example, have been shown to perform much worse on languages with elaborate morphology due to data sparsity problems \cite{khudanpur2006multilingual,bender2011achieving,Gerz:2018emnlp}. Such models were nevertheless more commonly used than more linguistically informed alternatives such as factored language models \cite{bilmes2003factored} that represent words as sets of features. Word embeddings have been widely used, in part because pre-trained embeddings covered a large part of the English vocabulary. 
However, word embeddings are not useful for tasks that require access to morphemes, e.g., semantic tasks in morphologically rich languages 
\cite{avraham-goldberg-2017-interplay}. 

While studies have demonstrated the ability of word embeddings to capture linguistic information in English, it remains unclear whether they capture the information needed for processing morphologically rich languages \cite{tsarfaty-etal-2020-spmrl}. A bias towards morphologically rich languages is also apparent in our tokenization algorithms. Subword tokenization performs poorly on languages with reduplication \cite{vania-lopez-2017-characters}, while byte pair encoding does not align well with morphology \cite{bostrom-durrett-2020-byte}. Consequently, languages with productive morphological systems also are disadvantaged when shared `language-universal' tokenizers are used in current large-scale multilingual language models \cite{acs:2019,Rust2020how} without any further vocabulary adaptation \cite{Wang20ExtendmBERT,Pfeiffer:2021emnlp}.

Another bias in our models relates to \textbf{word order}. In order for n-gram models to capture inter-word dependencies, words need to appear in the n-gram window. This will occur more frequently in languages with relatively fixed word order compared to languages with relatively free word order \cite{bender2011achieving}. Word embedding approaches such as skip-gram \cite{Mikolov2013skipgram} adhere to the same window-based approach and thus have similar weaknesses for languages with relatively free word order. LSTMs are also sensitive to word order and perform worse on agreement prediction in Basque, which is both morphologically richer and has a relatively free word order \cite{ravfogel-etal-2018-lstm} compared to English \cite{linzen2016assessing}. They have also been shown to transfer worse to distant languages for dependency parsing compared to self-attention models \cite{ahmad-etal-2019-difficulties}. Such biases concerning word order are not only inherent in our models but also in our algorithms. A recent unsupervised parsing algorithm \cite{Shen2018} has been shown to be biased towards right-branching structures and consequently performs better in right-branching languages like English \cite{dyer2019critical}. While the recent generation of self-attention based architectures can be seen as inherently order-agnostic, recent methods focusing on making attention more efficient \cite{tay2020efficient} introduce new biases into the models. Specifically, models that reduce the global attention to a local sliding window around the token \cite{Liu2018generating,child2019generating,zaheer2020big} may incur similar limitations as their n-gram and word embedding-based predecessors, performing worse on languages with relatively free word order.\footnote{An older work of \citet{khudanpur2006multilingual} argues that free word order is less of a problem as local order within phrases is relatively stable. However, it remains to be seen to what degree this affects current models.}

The singular focus on maximizing a performance metric such as accuracy introduces a bias towards models that are expressive enough to fit a given distribution well. Such models are typically \textbf{black-box} and learn highly non-linear relations that are generally not interpretable. Interpretability is generally studied in papers focusing exclusively on this topic; a recent example is BERTology \cite{rogers2020primer}. Studies proposing more interpretable methods typically build on state-of-the-art methods \cite{Weiss2018} and much work focuses on leveraging components such as attention for interpretability, which have not been designed with that goal in mind \cite{serrano-smith-2019-attention,wiegreffe-pinter-2019-attention}. As a result, researchers eschew directions focusing on models that are intrinsically more interpretable such as generalized additive models \cite{hastie2017generalized} and their extensions \cite{chang2021interpretable,Agarwal2021} but which have so far not been shown to match the performance of state-of-the-art methods.

As most datasets on which models are evaluated focus on sentences or short documents, state-of-the-art methods restrict their input size to around 512 tokens \cite{Devlin2019} and leverage methods that are \textbf{inefficient} when scaling to longer documents. This has led to the emergence of a wide range of more efficient models \cite{tay2020efficient}, which, however, are rarely used as baseline methods in NLP. Similarly, the standard pretrain-fine-tune paradigm \cite{ruder2019transfer} requires separate model copies to be stored for each task, and thus restricts work on multi-domain, multi-task, multi-lingual, multi-subpopulation methods that is enabled by more efficient and less resource-intensive \cite{Schwartz:2020cacm} fine-tuning methods \cite{houlsby2019parameter,pfeiffer-etal-2020-mad}


In sum, (what we typically consider as) standard baselines and state-of-the-art architectures favor languages with some characteristics over others and are optimized only for performance, which in turn propagates the \bias: If researchers study aspects such as multilinguality, efficiency, fairness or interpretability, they are likely to do so {\em with and for commonly used architectures} (i.e., often termed `standard architectures'), in order to reduce (too) many degrees of freedom in their empirical research. This is in many ways a sensible choice in order to maximize perceived relevance---and thereby, impact. However, as a result, multilinguality, efficiency, fairness, interpretability, and other research areas {\em inherit the same biases}, which typically slip under the radar. 

\rparagraph{Annotation Biases} Many NLP tasks can be cast differently and formulated in multiple ways, and differences may result in different annotation styles. Sentiment, for example, can be annotated at the document, sentence or word level \cite{socher-etal-2013-recursive}. In machine comprehension, answers are sometimes assumed to be continuous, but \citet{zhu-etal-2020-question} annotate discontinuous spans. In dependency parsing, different annotation guidelines can lead to very different downstream performance \cite{elming-etal-2013-stream}. How we annotate for a task may interact in complex ways with dimensions such as multilinguality, efficiency, fairness, and interpretability. The Universal Dependencies project \cite{nivre-etal-2020-universal} is motivated by the observation that not all dependency formalisms are easily applicable to all languages.  Aligning guidelines across languages has enabled researchers to ask interesting questions, but such attempts may limit the analysis of outlier languages \cite{DBLP:conf/tlt/CroftNLR17}. 

Other examples of annotation guidelines interacting with the above dimensions exist: Slight nuances in how annotation guidelines are formulated can lead to severe model biases \cite{hansen-sogaard-2021-guideline} and hurt model fairness. In interpretability, we can use feature attribution methods and word-level annotations to evaluate interpretability methods applied to sequence classifiers \cite{rei-sogaard-2018-zero}, but we cannot directly use feature attribution methods to obtain rationales for sequence labelers. Annotation biases can also stem from the characteristics of the annotators, including their domain experience \cite{10.1145/2488388.2488466}, demographics \cite{jorgensen-sogaard-2021-evaluation}, or educational level \cite{al-kuwatly-etal-2020-identifying}.  


Annotation biases form an integral part of the \bias: In NLP experiments, we commonly rely on the same pools of annotators, e.g., computer science students, professional linguists, or MTurk contributors. Sometimes these biases percolate through reuse of resources, e.g., through human or machine translation into new languages. Examples of such recycled resources include the ones introduced by \citet{xnli} and  \citet{kassner-etal-2021-multilingual}, among others. Even when such translation-based resources resonate with syntax and semantics of the target language, and are fluent and natural, they still suffer from \textit{translation artefacts}: they are often target-language surface realizations of source-language-based conceptual thinking \cite{Majewska:2022cod}. As a consequence, evaluations of cross-lingual transfer models on such data typically overestimate their performance as properties such as word order and even the choice of lexical units are inherently biased by the source language \cite{vanmassenhove-etal-2021-machine}. Put simply, the choice of the data creation protocol, e.g., translation-based versus data collection directly in the target language \cite{clark-etal-2020-tydi} can yield profound differences in model performance for some groups, or may have serious impact on the interpretability or computational efficiency (e.g., sample efficiency) of our models. 


\rparagraph{Selection Biases} For many years, the English Penn Treebank \cite{marcus-etal-1994-penn} was an integral part of the \point of NLP. This corpus consists entirely of newswire, i.e., articles and editorials from the Wall Street Journal, and arguably amplified the (existing) bias toward news articles. Since news articles tend to reflect a particular set of linguistic conventions, have a certain length, and are written by certain demographics, the bias toward news articles had an impact on the linguistic phenomena studied in NLP \cite{judge-etal-2006-questionbank}, led to under-representation of challenges with handling longer documents \cite{beltagy-etal-2021-beyond}, and had impact on early papers in fairness \cite{hovy-sogaard-2015-tagging}. Note how such a bias may interact in non-linear ways with efficiency, i.e., efficient methods for shorter documents need not be efficient for longer ones, or fairness, i.e., what mitigates gender biases in news articles need not mitigate gender biases in product reviews. 


\rparagraph{Protocol Biases} In the prototypical NLP experiment, the dataset is in the English language. As a consequence, it is also standard protocol in multilingual NLP to use English as a source language in zero-shot cross-lingual transfer \cite{Hu2020xtreme}. In practice, there are generally better source languages than English \cite{Ponti:2018acl,Lin2019,Turc2021}, and results are heavily biased by the common choice of English. For instance, effectiveness and efficiency of few-shot learning can be impacted by the choice of the source language \cite{Pfeiffer:2021emnlp,Zhao:2021acl}. English also dominates language pairs in machine translation, leading to lower performance for non-English translation directions \cite{Fan2020}, which are particularly important in multilingual societies. Again, such biases may interact in non-trivial ways with dimensions explored in NLP research: It is not inconceivable that there is an algorithm $A$ that is more fair, interpretable or efficient than algorithm $B$ on, say,  English-to-Czech transfer or translation, but not on German-to-Czech or French-to-Czech. 


\rparagraph{Organizational Biases} The above architectural, annotation, selection and protocol biases follow from the \bias, but they also conserve the \point. If our go-to architectures, resources, and experimental setups are tailored to some languages over others, some objectives over others, and some research paradigms over others, it is considerably more work to explore new sets of languages, new objectives, or new protocols. The organizational biases we discuss below may also reinforce the \bias.

The organization of our conferences and reviewing processes perpetuates certain biases. In particular, both during reviewing and for later presentation at conferences, papers are organized in areas. Upon submission, a paper is assigned to a single area. Reviewers are recruited for their expertise in a specific area, which they are associated with. Such a reviewing system incentivizes papers that make contributions to the chosen area, in order to appeal to the reviewers of this area and implicitly penalizes papers that make contributions along multiple dimensions, as reviewers unfamiliar with the related areas may not appreciate their inter-disciplinary or inter-areal magnitude or value. Even new initiatives that seek to improve reviewing such as ARR\footnote{\url{aclrollingreview.org/}} adhere to this area structure\footnote{\url{www.2022.aclweb.org/callpapers}} and thus further the \bias. A reviewing system that allows papers to be associated with multiple dimensions of research and that assigns reviewers with \emph{complementary} expertise---similar to TACL\footnote{\url{transacl.org/index.php/tacl}}---would ameliorate this situation. Once a paper is accepted, presentations at conferences are organized by areas, limiting audiences in most cases to members of said area and thereby reducing the cross-pollination of ideas.\footnote{Another previously pervasive organizational bias, which is now fortunately being institutionally mitigated within the *ACL community through dedicated mentoring programs and improved reviewing guidelines, concerned penalizing research papers for their non-native writing style, where it was frequently suggested to the authors whose native language is not English to `have their paper proofread by a native speaker'. As one hidden consequence, this attitude might have set a higher bar for the native speakers of minor and endangered languages working on such languages to put their research problems in the spotlight, that way also implicitly hindering more work of the entire community on these languages.}


\rparagraph{Unexplored Areas of the Research Manifold} The discussed biases, which seem to originate from the \bias, leave areas of the research manifold unexplored. Character-based language models are often reported to perform well for morphologically rich languages or on non-canonical text \cite{ma-etal-2020-charbert}, but little is known about their fairness properties, and attribution-based interpretability methods have not been developed for such models. Annotation biases that stem from annotator demographics have been studied for English POS tagging \cite{hovy-sogaard-2015-tagging} or English summarization \cite{jorgensen-sogaard-2021-evaluation}, for example, but there has been very little research on such biases for other languages. While linguistic differences among genders is shared among some languages, genders differ in very different ways between other languages, e.g., Spanish and Swedish \cite{johannsen-etal-2015-cross}. We discuss important unexplored areas of the research manifold in \S\ref{sec:blind_spots}, but first we briefly survey existing, multi-dimensional work, i.e., the counter-examples to our claim that NLP research is biased to one-dimensional extensions of the \pointlower. 

\section{Counter-Examples}

Most of the exceptions to our thesis about the `one-dimensionality' of NLP research, in our classification of ACL 2021 Oral Papers, came from studies of {\bf efficiency in a multilingual context}. Another example of this is \citet{ahia2021the}, who show that for low-resource languages, weight pruning hurts performance on tail phenomena, but improves robustness to out-of-distribution shifts---this is not observed in the \point (high-resource) regime. There are also studies of {\bf fairness in a multilingual context}. \citet{huang-etal-2020-multilingual}, for example, show significant differences in social bias for multilingual hate speech systems across different languages. \citet{zhao-etal-2020-gender} study gender bias in multilingual word embeddings and cross-lingual transfer. \citet{gonzalez-etal-2020-type} also study gender bias, but by relying on reflexive pronominal constructions that do not exist in the English language; this is a good example of research that would not have been possible taking \point as our point of departure. \citet{dayanik-pado-2021-disentangling} study adversarial debiasing in the context of a multilingual corpus and show some mitigation methods are more effective for some languages rather than others. \citet{Nozza2021} studies multilingual toxicity classification and finds that models misinterpret non-hateful language-specific taboo interjections as hate speech in some languages. There has been much less work on other combinations of these dimensions, e.g., {\bf fairness and efficiency}. \citet{hansen-sogaard-2021-lottery} show that weight pruning has disparate effects on performance across demographics and that the min-max difference in group disparities is negatively correlated with model size. \citet{renduchintala-etal-2021-gender} observe that techniques to make inference more efficient, e.g., greedy search, quantization, or shallow decoder models, have a small impact on performance, but dramatically amplify gender bias. In a rare study of {\bf fairness and interpretability}, \citet{vig2020investigating} propose a methodology to interpret which parts of a model are causally implicated in its behavior. They apply this methodology to analyze gender bias in pre-trained Transformers, finding that gender bias effects are sparse and concentrated in small parts of the network.

\section{Blind Spots} \label{sec:blind_spots}

We identified several under-explored areas on the research manifold. The common theme is a lack of studies of how dimensions such as multilinguality, fairness, efficiency, and interpretability interact. We now summarize some open problems that we believe are particularly important to address: (i) While recent work has begun to study the trade-off between {\bf efficiency and fairness}, this interaction remains largely unexplored, especially outside of the empirical risk minimization regime; (ii) {\bf fairness and interpretability} interact in potentially many ways, i.e., interpretability techniques may affect the fairness of the underlying models \cite{1548834}, but rationales may also, for example, be biased toward certain demographics in how they are presented \cite{feng2019ai,gonzalez-etal-2021-interaction}; (iii) finally, {\bf multilinguality and interpretability} seem heavily underexplored. While there exists resources for English for evaluating interpretability methods against gold-standard human annotations, there are, to the best of our knowledge, no such resources for other languages.\footnote{We again note that there are other possible dimensions, not studied in this work, that can expose more blind spots: e.g., {\bf fairness and multi-modality}, {\bf multilinguality and privacy}.}



\section{Contributing Factors}

We finally highlight possible factors that may contribute to the \bias.

\rparagraph{Biases in NLP Education} We hypothesize that early exposure to predominantly English-centric experiment settings and tasks using a single performance metric may potentially propagate further to more advanced NLP research. To investigate to what extent this may be the case, we created a short questionnaire, which we sent to a geographically diverse set of teachers, including first authors from the last Teaching NLP workshop \cite{teachingnlp-2021-teaching}, asking about the first experiment that they presented in their NLP 101 course. We received 71 responses in total. Our first question was: {\em The last time you taught an introductory NLP course, what was the first task you introduced the students to, or that they had to implement a model for?} The relative majority of respondents (31.9\%) said {\em sentiment analysis}, while 10.1\% indicated topic classification.\footnote{The remaining responses included NER, language modeling, language identification, hate speech detection, etc.} More importantly, we also asked them about the language of the data used in the experiment, and what metric they optimized for. More than three quarters of respondents reported that they used {\em English}~language training and evaluation data and more than three quarters of the respondents asked the students to optimize for {\em accuracy or F1}.  The choice of using English language datasets is particularly interesting in contrast to the native languages of the teachers and their students: In around two thirds of the classes, most students shared an L1 language that was not English; and less than a quarter of the teachers were L1 English speakers themselves. 
We extend this analysis to prototypical NLP experiments in undergraduate and graduate research based on five exemplary NLP textbooks, spanning 20 years (see Table~\ref{tab:textbooks}). We observe that they, like the teachers in our survey, take the same point of departure: an English-language experiment where we use supervised learning techniques to optimize for a standard performance metric, e.g., perplexity or error. We note an important difference, however: While the first four books largely ignore issues relating to fairness, interpretability, and efficiency, the most recent NLP textbook in Table~\ref{tab:textbooks} \cite{eisenstein2019introduction} discusses efficiency (briefly) and fairness (more thoroughly). Overall, we believe that teachers and educational materials should engage as early as possible with the multiple dimensions of NLP in order to sensitize researchers regarding these topics at the start of their careers.

\begin{table}[]
    \centering\scriptsize 
    \begin{tabular}{llll}
    \toprule {\bf Year}&{\bf Book}&{\bf Language}&{\bf Task}\\
    \midrule 
        1999&\citet{manning99foundations}&English-French&Alignment\\
         2009&\citet{jurafsky2009speech}&English&LM\\
         2009&\citet{BirdKleinLoper09}&English&Name cl.\\
         2013&\citet{9ad07f2f0c9f46149a59e72e06bf2e5b}&English&Doc.cl.\\
         2019&\citet{eisenstein2019introduction}&English&Doc.cl.\\
         \bottomrule 
    \end{tabular}
    \caption{First experiments in NLP textbooks. The objective across all books is optimizing for performance (AER, perplexity, or accuracy), rather than fairness, interpretability or efficiency.}
    \label{tab:textbooks}
\end{table}

\rparagraph{Commercial Factors} For commercially focused NLP, there is an incentive to focus on settings with many users, such as major languages. Similarly, as long as users do not mind using highly accurate black-box systems, researchers working on real-world applications can often afford to ignore dimensions such as interpretability and fairness.

\rparagraph{Momentum of the Status Quo} The \point is well supported by existing infrastructure, resources, baselines, and experimental results. Any work that seeks to depart from the standard setting has to work harder, not only to build systems and resources in order to establish comparability with existing work but also needs to argue convincingly the importance of such work. We provide practical recommendations in the next section on how we can facilitate such research as a community.

\section{Discussion}


\noindent \textbf{Is \bias not the Flipside of Scientific Protocol?} One potential argument {\em for} a community-wide \bias is that when studying the impact of some technique $t$, say a novel regularization term, we want to compare some system with and without $t$, i.e., control for all other factors. To maximize impact and ease workload, it makes sense at first sight to stick to a system and experimental protocol that is familiar or well-studied. Always returning to the \point is a way to control for all other factors and relating new findings to known territory. The reason why this is {\em only seemingly a good idea}, however, is that the factors we study in NLP research, may be non-linearly related. The fact that $t$ makes for a positive net contribution under one set of circumstances, does not imply that it would do so under different circumstances. This is illustrated most clearly by the research surveyed in \S\ref{sec:examples}. Ideally, we thus want to study the impact of $t$ under as many circumstances as possible, but in the absence of resources to do so, it is a better (collective) search strategy to apply $t$ to a {\em random} set of circumstances (within the space of relevant circumstances, of course).

\rparagraph{Comment on Meta-Research} This paper can be seen in the line of other meta-research \cite{davis1971s,lakatos1976falsification,weber2006reach,bloom2020ideas} that seeks to analyze research practices and whether a scientific field is heading in the right direction. Within the NLP community, much of such recent discussion has focused on the nature of leaderboards and the practice of benchmarking \cite{Ethayarajh2020,Ma2021}.

\vspace{1.2mm}
\noindent \textbf{Should Each Paper Aim to Cover All Dimensions?} We believe that a researcher should aspire to cover as many dimensions as possible with their research. Considering the dimensions of research encourages us to think more holistically about our research and its final impact. It may also accelerate progress as follow-up work will already be able to build on the insights of multi-dimensional analyses of new methods. It will also promote the cross-pollination of ideas, which will no longer be confined to their own sub-areas. While such multi-dimensional research may be cumbersome at the moment, we believe with the proper incentives and support, we can make it much more accessible.

\rparagraph{Practical Recommendations} What can we do to incentivize and facilitate multi-dimensional research? \textbf{i)} Currently, most NLP models are evaluated by one or two performance metrics, but we believe 
dimensions such as fairness, efficiency, and interpretability need to become integral criteria for model evaluation, in line with recent proposals of more user-centric leaderboards \cite{Ethayarajh2020,Ma2021}.
This requires new tools, e.g., to evaluate 
environmental impact \cite{henderson2020towards}, as well as new benchmarks, e.g., to evaluate fairness 
\cite{wilds2021} or efficiency \cite{liu2021towards}.
\textbf{ii)} We believe separate conference tracks (areas) lead to unfortunate silo effects and inhibit multi-dimensional research. 
Rather, we imagine conference submissions could provide a checklist with dimensions along which they make contributions, similar to reproducibility checklist. 
Reviewers can be assigned based on their expertise corresponding to different dimensions. \textbf{iii)} 
Finally, we recommend awareness of research prototypes and encourage reviewers and chairs to prioritize research that departs from prototypes in {\em multiple} dimensions, in order to explore new areas of the research manifold. 

\section{Conclusion}

We identified the prototypical NLP experiment through annotation experiments and surveys. We highlighted the associated \bias, which encourages research to go beyond the prototype in a single dimension. We discussed the problems resulting from this bias, by studying the area statistics of a recent NLP conference as well as by discussing historic and recent examples. We finally pointed to under-explored research directions and made practical recommendations to inspire more multi-dimensional research in NLP.

\section*{Acknowledgments}

Ivan Vuli\'{c} is funded by the ERC PoC Grant MultiConvAI (no. 957356) and a research donation from Huawei. Anders S{\o}gaard is sponsored by the Innovation Fund Denmark and a Google Focused Research Award. We thank Jacob Eisenstein for valuable feedback on a draft of this paper and the suggestion of the term `square one'.


\bibliography{custom}

\begin{thebibliography}{108}
\expandafter\ifx\csname natexlab\endcsname\relax\def\natexlab#1{#1}\fi

\bibitem[{Abdullah et~al.(2021)Abdullah, Zaitova, Avgustinova, M{\"o}bius, and
  Klakow}]{abdullah2021familiar}
Badr Abdullah, Iuliia Zaitova, Tania Avgustinova, Bernd M{\"o}bius, and
  Dietrich Klakow. 2021.
\newblock \href {https://aclanthology.org/2021.blackboxnlp-1.32} {How familiar
  does that sound? {C}ross-lingual representational similarity analysis of
  acoustic word embeddings}.
\newblock In \emph{Proceedings of the Fourth BlackboxNLP Workshop on Analyzing
  and Interpreting Neural Networks for NLP}, pages 407--419.

\bibitem[{\'Acs(2019)}]{acs:2019}
Judit \'Acs. 2019.
\newblock \href
  {http://juditacs.github.io/2019/02/19/bert-tokenization-stats.html}
  {Exploring {BERT}'s {V}ocabulary}.
\newblock \emph{Blog Post}.

\bibitem[{Agarwal et~al.(2021)Agarwal, Melnick, Frosst, Zhang, Lengerich,
  Caruana, and Hinton}]{Agarwal2021}
Rishabh Agarwal, Levi Melnick, Nicholas Frosst, Xuezhou Zhang, Ben Lengerich,
  Rich Caruana, and Geoffrey Hinton. 2021.
\newblock \href {http://arxiv.org/abs/2004.13912} {Neural additive models:
  {I}nterpretable machine learning with neural nets}.
\newblock In \emph{Proceedings of NeurIPS 2021}.

\bibitem[{Agarwal(2021)}]{1548834}
Sushant Agarwal. 2021.
\newblock \href
  {https://crcs.seas.harvard.edu/files/crcs/files/ai4sg-21_paper_23.pdf}
  {Trade-offs between fairness and interpretability in machine learning}.
\newblock In \emph{Proceedings of the IJCAI 2021 Workshop on AI for Social
  Good}.

\bibitem[{Ahia et~al.(2021)Ahia, Kreutzer, and Hooker}]{ahia2021the}
Orevaoghene Ahia, Julia Kreutzer, and Sara Hooker. 2021.
\newblock \href {https://aclanthology.org/2021.findings-emnlp.282} {The
  low-resource double bind: An empirical study of pruning for low-resource
  machine translation}.
\newblock In \emph{Findings of the Association for Computational Linguistics:
  EMNLP 2021}, pages 3316--3333.

\bibitem[{Ahmad et~al.(2019)Ahmad, Zhang, Ma, Hovy, Chang, and
  Peng}]{ahmad-etal-2019-difficulties}
Wasi Ahmad, Zhisong Zhang, Xuezhe Ma, Eduard Hovy, Kai-Wei Chang, and Nanyun
  Peng. 2019.
\newblock \href {https://doi.org/10.18653/v1/N19-1253} {On difficulties of
  cross-lingual transfer with order differences: A case study on dependency
  parsing}.
\newblock In \emph{Proceedings of NAACL-HLT 2019}, pages 2440--2452.

\bibitem[{Al~Kuwatly et~al.(2020)Al~Kuwatly, Wich, and
  Groh}]{al-kuwatly-etal-2020-identifying}
Hala Al~Kuwatly, Maximilian Wich, and Georg Groh. 2020.
\newblock \href {https://doi.org/10.18653/v1/2020.alw-1.21} {Identifying and
  measuring annotator bias based on annotators{'} demographic characteristics}.
\newblock In \emph{Proceedings of the Fourth Workshop on Online Abuse and
  Harms}, pages 184--190.

\bibitem[{Avraham and Goldberg(2017)}]{avraham-goldberg-2017-interplay}
Oded Avraham and Yoav Goldberg. 2017.
\newblock \href {https://aclanthology.org/E17-2067} {The interplay of semantics
  and morphology in word embeddings}.
\newblock In \emph{Proceedings of EACL 2017}, pages 422--426.

\bibitem[{Baroni and Lenci(2010)}]{baroni-lenci-2010-distributional}
Marco Baroni and Alessandro Lenci. 2010.
\newblock \href {https://doi.org/10.1162/coli_a_00016} {Distributional memory:
  A general framework for corpus-based semantics}.
\newblock \emph{Computational Linguistics}, 36(4):673--721.

\bibitem[{Beltagy et~al.(2021)Beltagy, Cohan, Hajishirzi, Min, and
  Peters}]{beltagy-etal-2021-beyond}
Iz~Beltagy, Arman Cohan, Hannaneh Hajishirzi, Sewon Min, and Matthew~E. Peters.
  2021.
\newblock \href {https://aclanthology.org/2021.naacl-tutorials.5} {Beyond
  paragraphs: {NLP} for long sequences}.
\newblock In \emph{Proceedings of NAACL-HLT 2021: Tutorials}, pages 20--24.

\bibitem[{Bender(2011)}]{bender2011achieving}
Emily~M. Bender. 2011.
\newblock \href
  {https://journals.linguisticsociety.org/elanguage/lilt/article/view/2624.html}
  {On achieving and evaluating language-independence in {NLP}}.
\newblock \emph{Linguistic Issues in Language Technology}, 6(3):1--26.

\bibitem[{Berger et~al.(1996)Berger, Della~Pietra, and
  Della~Pietra}]{berger-etal-1996-maximum}
Adam~L. Berger, Stephen~A. Della~Pietra, and Vincent~J. Della~Pietra. 1996.
\newblock \href {https://aclanthology.org/J96-1002} {A maximum entropy approach
  to natural language processing}.
\newblock \emph{Computational Linguistics}, 22(1):39--71.

\bibitem[{Bilmes and Kirchhoff(2003)}]{bilmes2003factored}
Jeff Bilmes and Katrin Kirchhoff. 2003.
\newblock \href {https://aclanthology.org/N03-2002.pdf} {Factored language
  models and generalized parallel backoff}.
\newblock In \emph{Companion Volume of the Proceedings of HLT-NAACL 2003-Short
  Papers}, pages 4--6.

\bibitem[{Bird et~al.(2009)Bird, Klein, and Loper}]{BirdKleinLoper09}
Steven Bird, Ewan Klein, and Edward Loper. 2009.
\newblock \href {https://doi.org/http://my.safaribooksonline.com/9780596516499}
  {\emph{Natural Language Processing with Python: Analyzing Text with the
  Natural Language Toolkit}}.
\newblock O'Reilly, Beijing.

\bibitem[{Birhane et~al.(2021)Birhane, Kalluri, Card, Agnew, Dotan, and
  Bao}]{Birhane2021}
Abeba Birhane, Pratyusha Kalluri, Dallas Card, William Agnew, Ravit Dotan, and
  Michelle Bao. 2021.
\newblock \href {https://arxiv.org/abs/2106.15590} {The values encoded in
  machine learning research}.
\newblock \emph{CoRR}, abs/2106.15590.

\bibitem[{Blodgett et~al.(2020)Blodgett, Barocas, III, and
  Wallach}]{Blodgett2020}
Su~Lin Blodgett, Solon Barocas, Hal~Daum{\'{e}} III, and Hanna~M. Wallach.
  2020.
\newblock \href {https://arxiv.org/abs/2005.14050} {Language (technology) is
  power: {A} critical survey of "bias" in {NLP}}.
\newblock \emph{CoRR}, abs/2005.14050.

\bibitem[{Bloom et~al.(2020)Bloom, Jones, Van~Reenen, and
  Webb}]{bloom2020ideas}
Nicholas Bloom, Charles~I Jones, John Van~Reenen, and Michael Webb. 2020.
\newblock Are ideas getting harder to find?
\newblock \emph{American Economic Review}, 110(4):1104--44.

\bibitem[{Bostrom and Durrett(2020)}]{bostrom-durrett-2020-byte}
Kaj Bostrom and Greg Durrett. 2020.
\newblock \href {https://doi.org/10.18653/v1/2020.findings-emnlp.414} {Byte
  pair encoding is suboptimal for language model pretraining}.
\newblock In \emph{Findings of the Association for Computational Linguistics:
  EMNLP 2020}, pages 4617--4624.

\bibitem[{Carletta(1996)}]{carletta-1996-assessing}
Jean Carletta. 1996.
\newblock \href {https://aclanthology.org/J96-2004} {Assessing agreement on
  classification tasks: The kappa statistic}.
\newblock \emph{Computational Linguistics}, 22(2):249--254.

\bibitem[{Chang et~al.(2021)Chang, Tan, Lengerich, Goldenberg, and
  Caruana}]{chang2021interpretable}
Chun-Hao Chang, Sarah Tan, Ben Lengerich, Anna Goldenberg, and Rich Caruana.
  2021.
\newblock \href {https://arxiv.org/pdf/2006.06466.pdf} {How interpretable and
  trustworthy are {GAM}s?}
\newblock In \emph{Proceedings of KDD 2021}, pages 95--105.

\bibitem[{Child et~al.(2019)Child, Gray, Radford, and
  Sutskever}]{child2019generating}
Rewon Child, Scott Gray, Alec Radford, and Ilya Sutskever. 2019.
\newblock \href {http://arxiv.org/abs/1904.10509} {Generating long sequences
  with sparse {T}ransformers}.
\newblock \emph{CoRR}, abs/1904.10509.

\bibitem[{Clark et~al.(2020)Clark, Choi, Collins, Garrette, Kwiatkowski,
  Nikolaev, and Palomaki}]{clark-etal-2020-tydi}
Jonathan~H. Clark, Eunsol Choi, Michael Collins, Dan Garrette, Tom Kwiatkowski,
  Vitaly Nikolaev, and Jennimaria Palomaki. 2020.
\newblock \href {https://doi.org/10.1162/tacl_a_00317} {{T}y{D}i {QA}: A
  benchmark for information-seeking question answering in typologically diverse
  languages}.
\newblock \emph{Transactions of the Association for Computational Linguistics},
  8:454--470.

\bibitem[{Conneau et~al.(2018)Conneau, Rinott, Lample, Williams, Bowman,
  Schwenk, and Stoyanov}]{xnli}
Alexis Conneau, Ruty Rinott, Guillaume Lample, Adina Williams, Samuel Bowman,
  Holger Schwenk, and Veselin Stoyanov. 2018.
\newblock \href {https://aclanthology.org/D18-1269} {{XNLI: E}valuating
  cross-lingual sentence representations}.
\newblock In \emph{Proceedings of EMNLP 2018}, pages 2475--2485.

\bibitem[{Croft et~al.(2017)Croft, Nordquist, Looney, and
  Regan}]{DBLP:conf/tlt/CroftNLR17}
William Croft, Dawn Nordquist, Katherine Looney, and Michael Regan. 2017.
\newblock \href {http://ceur-ws.org/Vol-1779/05croft.pdf} {Linguistic typology
  meets universal dependencies}.
\newblock In \emph{Proceedings of the 15th International Workshop on Treebanks
  and Linguistic Theories (TLT15)}, pages 63--75.

\bibitem[{Das-Smaal(1990)}]{DASSMAAL1990349}
Edith~A. Das-Smaal. 1990.
\newblock \href {https://doi.org/https://doi.org/10.1016/S0166-4115(08)61332-1}
  {Biases in categorization}.
\newblock volume~68 of \emph{Advances in Psychology}, pages 349--386.
  North-Holland.

\bibitem[{Davis(1971)}]{davis1971s}
Murray~S Davis. 1971.
\newblock That's interesting! towards a phenomenology of sociology and a
  sociology of phenomenology.
\newblock \emph{Philosophy of the social sciences}, 1(2):309--344.

\bibitem[{Dayanik and Pad{\'o}(2021)}]{dayanik-pado-2021-disentangling}
Erenay Dayanik and Sebastian Pad{\'o}. 2021.
\newblock \href {https://aclanthology.org/2021.wassa-1.6} {Disentangling
  document topic and author gender in multiple languages: Lessons for
  adversarial debiasing}.
\newblock In \emph{Proceedings of the Eleventh Workshop on Computational
  Approaches to Subjectivity, Sentiment and Social Media Analysis}, pages
  50--61.

\bibitem[{Devlin et~al.(2019)Devlin, Chang, Lee, and Toutanova}]{Devlin2019}
Jacob Devlin, Ming-Wei Chang, Kenton Lee, and Kristina Toutanova. 2019.
\newblock \href {http://arxiv.org/abs/1810.04805} {{BERT: Pre-training of Deep
  Bidirectional Transformers for Language Understanding}}.
\newblock In \emph{Proceedings of NAACL-HLT 2019}.

\bibitem[{Dyer et~al.(2019)Dyer, Melis, and Blunsom}]{dyer2019critical}
Chris Dyer, G{\'{a}}bor Melis, and Phil Blunsom. 2019.
\newblock \href {http://arxiv.org/abs/1909.09428} {A critical analysis of
  biased parsers in unsupervised parsing}.
\newblock \emph{CoRR}, abs/1909.09428.

\bibitem[{Eisenstein(2019)}]{eisenstein2019introduction}
Jacob Eisenstein. 2019.
\newblock \href {https://books.google.se/books?id=72yuDwAAQBAJ}
  {\emph{Introduction to Natural Language Processing}}.
\newblock Adaptive Computation and Machine Learning series. MIT Press.

\bibitem[{Elming et~al.(2013)Elming, Johannsen, Klerke, Lapponi,
  Martinez~Alonso, and S{\o}gaard}]{elming-etal-2013-stream}
Jakob Elming, Anders Johannsen, Sigrid Klerke, Emanuele Lapponi, Hector
  Martinez~Alonso, and Anders S{\o}gaard. 2013.
\newblock \href {https://aclanthology.org/N13-1070} {Down-stream effects of
  tree-to-dependency conversions}.
\newblock In \emph{Proceedings of NAACL-HLT 2013}, pages 617--626.

\bibitem[{Ethayarajh and Jurafsky(2020)}]{Ethayarajh2020}
Kawin Ethayarajh and Dan Jurafsky. 2020.
\newblock \href {https://aclanthology.org/2020.emnlp-main.393} {Utility is in
  the eye of the user: A critique of {NLP} leaderboards}.
\newblock In \emph{Proceedings of EMNLP 2020}, pages 4846--4853.

\bibitem[{Fan et~al.(2020)Fan, Bhosale, Schwenk, Ma, El-Kishky, Goyal, Baines,
  Celebi, Wenzek, Chaudhary, Goyal, Birch, Liptchinsky, Edunov, Grave, Auli,
  and Joulin}]{Fan2020}
Angela Fan, Shruti Bhosale, Holger Schwenk, Zhiyi Ma, Ahmed El-Kishky,
  Siddharth Goyal, Mandeep Baines, Onur Celebi, Guillaume Wenzek, Vishrav
  Chaudhary, Naman Goyal, Tom Birch, Vitaliy Liptchinsky, Sergey Edunov,
  Edouard Grave, Michael Auli, and Armand Joulin. 2020.
\newblock \href {http://arxiv.org/abs/2010.11125} {{Beyond English-Centric
  Multilingual Machine Translation}}.
\newblock \emph{arXiv preprint arXiv:2010.11125}.

\bibitem[{Feng and Boyd{-}Graber(2018)}]{feng2019ai}
Shi Feng and Jordan~L. Boyd{-}Graber. 2018.
\newblock \href {http://arxiv.org/abs/1810.09648} {What can {AI} do for me:
  Evaluating machine learning interpretations in cooperative play}.
\newblock \emph{CoRR}, abs/1810.09648.

\bibitem[{Gerz et~al.(2018)Gerz, Vuli{\'c}, Ponti, Reichart, and
  Korhonen}]{Gerz:2018emnlp}
Daniela Gerz, Ivan Vuli{\'c}, Edoardo~Maria Ponti, Roi Reichart, and Anna
  Korhonen. 2018.
\newblock \href {https://www.aclweb.org/anthology/D18-1029} {On the relation
  between linguistic typology and (limitations of) multilingual language
  modeling}.
\newblock In \emph{Proceedings of EMNLP 2018}, pages 316--327.

\bibitem[{Gonz{\'a}lez et~al.(2020)Gonz{\'a}lez, Barrett, Hvingelby, Webster,
  and S{\o}gaard}]{gonzalez-etal-2020-type}
Ana~Valeria Gonz{\'a}lez, Maria Barrett, Rasmus Hvingelby, Kellie Webster, and
  Anders S{\o}gaard. 2020.
\newblock \href {https://aclanthology.org/2020.emnlp-main.209} {Type {B}
  reflexivization as an unambiguous testbed for multilingual multi-task gender
  bias}.
\newblock In \emph{Proceedings of EMNLP 2020}, pages 2637--2648.

\bibitem[{Gonz{\'a}lez et~al.(2021)Gonz{\'a}lez, Rogers, and
  S{\o}gaard}]{gonzalez-etal-2021-interaction}
Ana~Valeria Gonz{\'a}lez, Anna Rogers, and Anders S{\o}gaard. 2021.
\newblock \href {https://doi.org/10.18653/v1/2021.findings-acl.259} {On the
  interaction of belief bias and explanations}.
\newblock In \emph{Findings of the Association for Computational Linguistics:
  ACL-IJCNLP 2021}, pages 2930--2942.

\bibitem[{Grosz et~al.(1995)Grosz, Joshi, and
  Weinstein}]{grosz-etal-1995-centering}
Barbara~J. Grosz, Aravind~K. Joshi, and Scott Weinstein. 1995.
\newblock \href {https://aclanthology.org/J95-2003} {{C}entering: A framework
  for modeling the local coherence of discourse}.
\newblock \emph{Computational Linguistics}, 21(2):203--225.

\bibitem[{Hansen and
  S{\o}gaard(2021{\natexlab{a}})}]{hansen-sogaard-2021-guideline}
Victor Petr{\'e}n~Bach Hansen and Anders S{\o}gaard. 2021{\natexlab{a}}.
\newblock \href {https://doi.org/10.18653/v1/2021.bppf-1.2} {Guideline bias in
  {W}izard-of-{O}z dialogues}.
\newblock In \emph{Proceedings of the 1st Workshop on Benchmarking: Past,
  Present and Future}, pages 8--14.

\bibitem[{Hansen and
  S{\o}gaard(2021{\natexlab{b}})}]{hansen-sogaard-2021-lottery}
Victor Petr{\'e}n~Bach Hansen and Anders S{\o}gaard. 2021{\natexlab{b}}.
\newblock \href {https://doi.org/10.18653/v1/2021.findings-acl.284} {Is the
  lottery fair? evaluating winning tickets across demographics}.
\newblock In \emph{Findings of the Association for Computational Linguistics:
  ACL-IJCNLP 2021}, pages 3214--3224.

\bibitem[{Hastie and Tibshirani(2017)}]{hastie2017generalized}
Trevor~J. Hastie and Robert~J. Tibshirani. 2017.
\newblock \emph{Generalized additive models}.
\newblock Routledge.

\bibitem[{Henderson et~al.(2020)Henderson, Hu, Romoff, Brunskill, Jurafsky, and
  Pineau}]{henderson2020towards}
Peter Henderson, Jieru Hu, Joshua Romoff, Emma Brunskill, Dan Jurafsky, and
  Joelle Pineau. 2020.
\newblock \href {https://arxiv.org/pdf/2002.05651.pdf} {Towards the systematic
  reporting of the energy and carbon footprints of machine learning}.
\newblock \emph{Journal of Machine Learning Research}, 21(248):1--43.

\bibitem[{Houlsby et~al.(2019)Houlsby, Giurgiu, Jastrzebski, Morrone,
  De~Laroussilhe, Gesmundo, Attariyan, and Gelly}]{houlsby2019parameter}
Neil Houlsby, Andrei Giurgiu, Stanislaw Jastrzebski, Bruna Morrone, Quentin
  De~Laroussilhe, Andrea Gesmundo, Mona Attariyan, and Sylvain Gelly. 2019.
\newblock \href {https://arxiv.org/abs/1902.00751} {Parameter-efficient
  transfer learning for {NLP}}.
\newblock In \emph{Proceedings of ICML 2019}, pages 2790--2799.

\bibitem[{Hovy and S{\o}gaard(2015)}]{hovy-sogaard-2015-tagging}
Dirk Hovy and Anders S{\o}gaard. 2015.
\newblock \href {https://aclanthology.org/P15-2079} {Tagging performance
  correlates with author age}.
\newblock In \emph{Proceedings of ACL-IJCNLP 2015}, pages 483--488.

\bibitem[{Hu et~al.(2020)Hu, Ruder, Siddhant, Neubig, Firat, and
  Johnson}]{Hu2020xtreme}
Junjie Hu, Sebastian Ruder, Aditya Siddhant, Graham Neubig, Orhan Firat, and
  Melvin Johnson. 2020.
\newblock \href {http://arxiv.org/abs/arXiv:2003.11080v1} {{XTREME: A Massively
  Multilingual Multi-task Benchmark for Evaluating Cross-lingual
  Generalization}}.
\newblock In \emph{Proceedings of ICML 2020}.

\bibitem[{Huang et~al.(2020)Huang, Xing, Dernoncourt, and
  Paul}]{huang-etal-2020-multilingual}
Xiaolei Huang, Linzi Xing, Franck Dernoncourt, and Michael~J. Paul. 2020.
\newblock \href {https://aclanthology.org/2020.lrec-1.180} {Multilingual
  {T}witter corpus and baselines for evaluating demographic bias in hate speech
  recognition}.
\newblock In \emph{Proceedings of LREC 2020}, pages 1440--1448.

\bibitem[{Johannsen et~al.(2015)Johannsen, Hovy, and
  S{\o}gaard}]{johannsen-etal-2015-cross}
Anders Johannsen, Dirk Hovy, and Anders S{\o}gaard. 2015.
\newblock \href {https://doi.org/10.18653/v1/K15-1011} {Cross-lingual syntactic
  variation over age and gender}.
\newblock In \emph{Proceedings of CoNLL 2015}, pages 103--112.

\bibitem[{J{\o}rgensen and
  S{\o}gaard(2021)}]{jorgensen-sogaard-2021-evaluation}
Anna J{\o}rgensen and Anders S{\o}gaard. 2021.
\newblock \href {https://aclanthology.org/2021.newsum-1.6} {Evaluation of
  summarization systems across gender, age, and race}.
\newblock In \emph{Proceedings of the Third Workshop on New Frontiers in
  Summarization}, pages 51--56.

\bibitem[{Joshi et~al.(2020)Joshi, Santy, Budhiraja, Bali, and
  Choudhury}]{Joshi2020}
Pratik Joshi, Sebastin Santy, Amar Budhiraja, Kalika Bali, and Monojit
  Choudhury. 2020.
\newblock \href {http://arxiv.org/abs/2004.09095} {{The State and Fate of
  Linguistic Diversity and Inclusion in the NLP World}}.
\newblock In \emph{Proceedings of ACL 2020}.

\bibitem[{Judge et~al.(2006)Judge, Cahill, and van
  Genabith}]{judge-etal-2006-questionbank}
John Judge, Aoife Cahill, and Josef van Genabith. 2006.
\newblock \href {https://doi.org/10.3115/1220175.1220238} {{Q}uestion{B}ank:
  Creating a corpus of parse-annotated questions}.
\newblock In \emph{Proceedings of the 21st International Conference on
  Computational Linguistics and 44th Annual Meeting of the Association for
  Computational Linguistics}, pages 497--504.

\bibitem[{Jurafsky and Martin(2009)}]{jurafsky2009speech}
Dan Jurafsky and James~H. Martin. 2009.
\newblock \href
  {http://www.amazon.com/Speech-Language-Processing-2nd-Edition/dp/0131873210/ref=pd_bxgy_b_img_y}
  {\emph{Speech and language processing : an introduction to natural language
  processing, computational linguistics, and speech recognition}}.
\newblock Pearson Prentice Hall, Upper Saddle River, N.J.

\bibitem[{Jurgens et~al.(2021)Jurgens, Kolhatkar, Li, Mieskes, and
  Pedersen}]{teachingnlp-2021-teaching}
David Jurgens, Varada Kolhatkar, Lucy Li, Margot Mieskes, and Ted Pedersen,
  editors. 2021.
\newblock \href {https://aclanthology.org/2021.teachingnlp-1.0}
  {\emph{Proceedings of the Fifth Workshop on Teaching NLP}}.

\bibitem[{Kassner et~al.(2021)Kassner, Dufter, and
  Sch{\"u}tze}]{kassner-etal-2021-multilingual}
Nora Kassner, Philipp Dufter, and Hinrich Sch{\"u}tze. 2021.
\newblock \href {https://doi.org/10.18653/v1/2021.eacl-main.284} {Multilingual
  {LAMA}: Investigating knowledge in multilingual pretrained language models}.
\newblock In \emph{Proceedings of EACL 2021}, pages 3250--3258.

\bibitem[{Khudanpur(2006)}]{khudanpur2006multilingual}
Sanjeev~P Khudanpur. 2006.
\newblock Multilingual language modeling.
\newblock \emph{Multilingual Speech Processing}, page 169.

\bibitem[{Koh et~al.(2021)Koh, Sagawa, Marklund, Xie, Zhang, Balsubramani, Hu,
  Yasunaga, Phillips, Gao, Lee, David, Stavness, Guo, Earnshaw, Haque, Beery,
  Leskovec, Kundaje, Pierson, Levine, Finn, and Liang}]{wilds2021}
Pang~Wei Koh, Shiori Sagawa, Henrik Marklund, Sang~Michael Xie, Marvin Zhang,
  Akshay Balsubramani, Weihua Hu, Michihiro Yasunaga, Richard~Lanas Phillips,
  Irena Gao, Tony Lee, Etienne David, Ian Stavness, Wei Guo, Berton~A.
  Earnshaw, Imran~S. Haque, Sara Beery, Jure Leskovec, Anshul Kundaje, Emma
  Pierson, Sergey Levine, Chelsea Finn, and Percy Liang. 2021.
\newblock \href {https://arxiv.org/abs/2012.07421} {{WILDS}: A benchmark of
  in-the-wild distribution shifts}.
\newblock In \emph{Proceedings of ICML 2021}.

\bibitem[{Lakatos(1976)}]{lakatos1976falsification}
Imre Lakatos. 1976.
\newblock Falsification and the methodology of scientific research programmes.
\newblock In \emph{Can theories be refuted?}, pages 205--259. Springer.

\bibitem[{Lin et~al.(2019)Lin, Chen, Lee, Li, Zhang, Xia, Rijhwani, He, Zhang,
  Ma, Anastasopoulos, Littell, and Neubig}]{Lin2019}
Yu-Hsiang Lin, Chian-Yu Chen, Jean Lee, Zirui Li, Yuyan Zhang, Mengzhou Xia,
  Shruti Rijhwani, Junxian He, Zhisong Zhang, Xuezhe Ma, Antonios
  Anastasopoulos, Patrick Littell, and Graham Neubig. 2019.
\newblock \href {http://arxiv.org/abs/1905.12688} {{Choosing Transfer Languages
  for Cross-Lingual Learning}}.
\newblock In \emph{Proceedings of ACL 2019}.

\bibitem[{Linzen et~al.(2016)Linzen, Dupoux, and
  Goldberg}]{linzen2016assessing}
Tal Linzen, Emmanuel Dupoux, and Yoav Goldberg. 2016.
\newblock \href {https://arxiv.org/abs/1611.01368} {Assessing the ability of
  {LSTM}s to learn syntax-sensitive dependencies}.
\newblock \emph{Transactions of the Association for Computational Linguistics},
  4:521--535.

\bibitem[{Liu et~al.(2021{\natexlab{a}})Liu, Bugliarello, Ponti, Reddy,
  Collier, and Elliott}]{Liu2021marvl}
Fangyu Liu, Emanuele Bugliarello, Edoardo~Maria Ponti, Siva Reddy, Nigel
  Collier, and Desmond Elliott. 2021{\natexlab{a}}.
\newblock \href {http://arxiv.org/abs/2109.13238} {{Visually Grounded Reasoning
  across Languages and Cultures}}.
\newblock In \emph{Proceedings of EMNLP 2021}.

\bibitem[{Liu et~al.(2018)Liu, Saleh, Pot, Goodrich, Sepassi, Kaiser, and
  Shazeer}]{Liu2018generating}
Peter~J. Liu, Mohammad Saleh, Etienne Pot, Ben Goodrich, Ryan Sepassi,
  {\L}ukasz Kaiser, and Noam Shazeer. 2018.
\newblock \href {http://arxiv.org/abs/arXiv:1801.10198v1} {{Generating
  Wikipedia by Summarizing Long Sequences}}.
\newblock In \emph{Proceedings of ICLR 2018}.

\bibitem[{Liu et~al.(2021{\natexlab{b}})Liu, Sun, He, Wu, Zhang, Jiang, Cao,
  Huang, and Qiu}]{liu2021towards}
Xiangyang Liu, Tianxiang Sun, Junliang He, Lingling Wu, Xinyu Zhang, Hao Jiang,
  Zhao Cao, Xuanjing Huang, and Xipeng Qiu. 2021{\natexlab{b}}.
\newblock Towards efficient nlp: A standard evaluation and a strong baseline.
\newblock \emph{arXiv preprint arXiv:2110.07038}.

\bibitem[{Ma et~al.(2020)Ma, Cui, Si, Liu, Wang, and
  Hu}]{ma-etal-2020-charbert}
Wentao Ma, Yiming Cui, Chenglei Si, Ting Liu, Shijin Wang, and Guoping Hu.
  2020.
\newblock \href {https://doi.org/10.18653/v1/2020.coling-main.4} {{C}har{BERT}:
  Character-aware pre-trained language model}.
\newblock In \emph{Proceedings of the 28th International Conference on
  Computational Linguistics}, pages 39--50, Barcelona, Spain (Online).
  International Committee on Computational Linguistics.

\bibitem[{Ma et~al.(2021)Ma, Ethayarajh, Thrush, Jain, Wu, Jia, Potts,
  Williams, and Kiela}]{Ma2021}
Zhiyi Ma, Kawin Ethayarajh, Tristan Thrush, Somya Jain, Ledell Wu, Robin Jia,
  Christopher Potts, Adina Williams, and Douwe Kiela. 2021.
\newblock \href {https://arxiv.org/abs/2106.06052} {Dynaboard: An
  evaluation-as-a-service platform for holistic next-generation benchmarking}.
\newblock \emph{CoRR}, abs/2106.06052.

\bibitem[{Majewska et~al.(2022)Majewska, Razumovskaia, Ponti, Vulic, and
  Korhonen}]{Majewska:2022cod}
Olga Majewska, Evgeniia Razumovskaia, Edoardo~Maria Ponti, Ivan Vulic, and Anna
  Korhonen. 2022.
\newblock \href {https://arxiv.org/abs/2201.13405} {Cross-lingual dialogue
  dataset creation via outline-based generation}.
\newblock \emph{CoRR}, abs/2201.13405.

\bibitem[{Manning and Sch{\"u}tze(1999)}]{manning99foundations}
Christopher~D. Manning and Hinrich Sch{\"u}tze. 1999.
\newblock \href {http://nlp.stanford.edu/fsnlp/} {\emph{Foundations of
  Statistical Natural Language Processing}}.
\newblock The {MIT} Press, Cambridge, Massachusetts.

\bibitem[{Marcus et~al.(1994)Marcus, Kim, Marcinkiewicz, MacIntyre, Bies,
  Ferguson, Katz, and Schasberger}]{marcus-etal-1994-penn}
Mitchell Marcus, Grace Kim, Mary~Ann Marcinkiewicz, Robert MacIntyre, Ann Bies,
  Mark Ferguson, Karen Katz, and Britta Schasberger. 1994.
\newblock \href {https://aclanthology.org/H94-1020} {The {P}enn {T}reebank:
  Annotating predicate argument structure}.
\newblock In \emph{{H}uman {L}anguage {T}echnology: Proceedings of a Workshop}.

\bibitem[{McAuley and Leskovec(2013)}]{10.1145/2488388.2488466}
Julian~John McAuley and Jure Leskovec. 2013.
\newblock \href {https://doi.org/10.1145/2488388.2488466} {From amateurs to
  connoisseurs: Modeling the evolution of user expertise through online
  reviews}.
\newblock In \emph{Proceedings of WWW 2013}, page 897–908.

\bibitem[{Mikolov et~al.(2013)Mikolov, Chen, Corrado, and
  Dean}]{Mikolov2013skipgram}
Tomas Mikolov, Kai Chen, Greg Corrado, and Jeffrey Dean. 2013.
\newblock \href {http://arxiv.org/abs/1310.4546} {{Distributed Representations
  of Words and Phrases and their Compositionality}}.
\newblock In \emph{Proceedings of NeurIPS 2013}.

\bibitem[{Nivre et~al.(2020)Nivre, de~Marneffe, Ginter, Haji{\v{c}}, Manning,
  Pyysalo, Schuster, Tyers, and Zeman}]{nivre-etal-2020-universal}
Joakim Nivre, Marie-Catherine de~Marneffe, Filip Ginter, Jan Haji{\v{c}},
  Christopher~D. Manning, Sampo Pyysalo, Sebastian Schuster, Francis Tyers, and
  Daniel Zeman. 2020.
\newblock \href {https://aclanthology.org/2020.lrec-1.497} {{U}niversal
  {D}ependencies v2: An evergrowing multilingual treebank collection}.
\newblock In \emph{Proceedings of LREC 2020}, pages 4034--4043.

\bibitem[{Nozza(2021)}]{Nozza2021}
Debora Nozza. 2021.
\newblock \href {https://aclanthology.org/2021.acl-short.114} {{Exposing the
  Limits of Zero-shot Cross-lingual Hate Speech Detection}}.
\newblock In \emph{Proceedings of ACL 2021}, pages 907--914.

\bibitem[{Ott et~al.(2011)Ott, Choi, Cardie, and
  Hancock}]{ott-etal-2011-finding}
Myle Ott, Yejin Choi, Claire Cardie, and Jeffrey~T. Hancock. 2011.
\newblock \href {https://aclanthology.org/P11-1032} {Finding deceptive opinion
  spam by any stretch of the imagination}.
\newblock In \emph{Proceedings of ACL 2011}, pages 309--319.

\bibitem[{Pfeiffer et~al.(2020)Pfeiffer, Vuli{\'c}, Gurevych, and
  Ruder}]{pfeiffer-etal-2020-mad}
Jonas Pfeiffer, Ivan Vuli{\'c}, Iryna Gurevych, and Sebastian Ruder. 2020.
\newblock \href {https://aclanthology.org/2020.emnlp-main.617} {{MAD-X}: {A}n
  {A}dapter-{B}ased {F}ramework for {M}ulti-{T}ask {C}ross-{L}ingual
  {T}ransfer}.
\newblock In \emph{Proceedings of EMNLP 2020}, pages 7654--7673.

\bibitem[{Pfeiffer et~al.(2021)Pfeiffer, Vuli\'{c}, Gurevych, and
  Ruder}]{Pfeiffer:2021emnlp}
Jonas Pfeiffer, Ivan Vuli\'{c}, Iryna Gurevych, and Sebastian Ruder. 2021.
\newblock \href {https://arxiv.org/abs/2012.15562} {{UNK}s everywhere:
  {A}dapting multilingual language models to new scripts}.
\newblock In \emph{Proceedings of EMNLP 2021}.

\bibitem[{Ponti et~al.(2018)Ponti, Reichart, Korhonen, and
  Vuli{\'c}}]{Ponti:2018acl}
Edoardo~Maria Ponti, Roi Reichart, Anna Korhonen, and Ivan Vuli{\'c}. 2018.
\newblock \href {https://doi.org/10.18653/v1/P18-1142} {Isomorphic transfer of
  syntactic structures in cross-lingual {NLP}}.
\newblock In \emph{Proceedings of ACL 2018}, pages 1531--1542.

\bibitem[{Ravfogel et~al.(2018)Ravfogel, Goldberg, and
  Tyers}]{ravfogel-etal-2018-lstm}
Shauli Ravfogel, Yoav Goldberg, and Francis Tyers. 2018.
\newblock \href {https://aclanthology.org/W18-5412} {Can {LSTM} learn to
  capture agreement? the case of {B}asque}.
\newblock In \emph{Proceedings of the 2018 {EMNLP} Workshop {B}lackbox{NLP}:
  Analyzing and Interpreting Neural Networks for {NLP}}, pages 98--107.

\bibitem[{Rei and S{\o}gaard(2018)}]{rei-sogaard-2018-zero}
Marek Rei and Anders S{\o}gaard. 2018.
\newblock \href {https://aclanthology.org/N18-1027} {Zero-shot sequence
  labeling: Transferring knowledge from sentences to tokens}.
\newblock In \emph{Proceedings of NAACL-HLT 2018}, pages 293--302.

\bibitem[{Renduchintala et~al.(2021)Renduchintala, Diaz, Heafield, Li, and
  Diab}]{renduchintala-etal-2021-gender}
Adithya Renduchintala, Denise Diaz, Kenneth Heafield, Xian Li, and Mona Diab.
  2021.
\newblock \href {https://aclanthology.org/2021.acl-short.15} {Gender bias
  amplification during speed-quality optimization in neural machine
  translation}.
\newblock In \emph{Proceedings of ACL-IJCNLP 2021}, pages 99--109.

\bibitem[{Rogers et~al.(2020)Rogers, Kovaleva, and
  Rumshisky}]{rogers2020primer}
Anna Rogers, Olga Kovaleva, and Anna Rumshisky. 2020.
\newblock \href {https://aclanthology.org/2020.tacl-1.54/} {A primer in
  {BERTology: Wh}at we know about how {BERT} works}.
\newblock \emph{Transactions of the Association for Computational Linguistics},
  8:842--866.

\bibitem[{Ruder et~al.(2019)Ruder, Peters, Swayamdipta, and
  Wolf}]{ruder2019transfer}
Sebastian Ruder, Matthew~E Peters, Swabha Swayamdipta, and Thomas Wolf. 2019.
\newblock Transfer learning in natural language processing.
\newblock In \emph{Proceedings of the 2019 Conference of the North American
  Chapter of the Association for Computational Linguistics: Tutorials}, pages
  15--18.

\bibitem[{Rust et~al.(2021)Rust, Pfeiffer, Vuli\'{c}, Ruder, and
  Gurevych}]{Rust2020how}
Phillip Rust, Jonas Pfeiffer, Ivan Vuli\'{c}, Sebastian Ruder, and Iryna
  Gurevych. 2021.
\newblock \href {https://doi.org/10.18653/v1/2021.acl-long.243} {{How Good is
  Your Tokenizer? {O}n the Monolingual Performance of Multilingual Language
  Models}}.
\newblock In \emph{Proceedings of ACL-IJCNLP 2021}, pages 3118--3135.

\bibitem[{Schwartz et~al.(2020)Schwartz, Dodge, Smith, and
  Etzioni}]{Schwartz:2020cacm}
Roy Schwartz, Jesse Dodge, Noah~A. Smith, and Oren Etzioni. 2020.
\newblock \href {https://doi.org/10.1145/3381831} {Green {AI}}.
\newblock \emph{Communications of the {ACM}}, 63(12):54--63.

\bibitem[{Serrano and Smith(2019)}]{serrano-smith-2019-attention}
Sofia Serrano and Noah~A. Smith. 2019.
\newblock \href {https://aclanthology.org/P19-1282} {Is attention
  interpretable?}
\newblock In \emph{Proceedings of ACL 2019}, pages 2931--2951.

\bibitem[{Shen et~al.(2018)Shen, Lin, Huang, and Courville}]{Shen2018}
Yikang Shen, Zhouhan Lin, Chin-wei Huang, and Aaron Courville. 2018.
\newblock \href {https://arxiv.org/pdf/1711.02013.pdf} {{Neural Language
  Modeling by Jointly Learning Syntax and Lexicon}}.
\newblock In \emph{Proceedings of ICLR 2018}.

\bibitem[{Sherman(1985)}]{sherman85}
Tracy Sherman. 1985.
\newblock Categorization skills in infants.
\newblock \emph{Child Development}, 56(6):1561--73.

\bibitem[{Socher et~al.(2013)Socher, Perelygin, Wu, Chuang, Manning, Ng, and
  Potts}]{socher-etal-2013-recursive}
Richard Socher, Alex Perelygin, Jean Wu, Jason Chuang, Christopher~D. Manning,
  Andrew Ng, and Christopher Potts. 2013.
\newblock \href {https://aclanthology.org/D13-1170} {Recursive deep models for
  semantic compositionality over a sentiment treebank}.
\newblock In \emph{Proceedings of EMNLP 2013}, pages 1631--1642.

\bibitem[{S{\o}gaard(2013)}]{9ad07f2f0c9f46149a59e72e06bf2e5b}
Anders S{\o}gaard. 2013.
\newblock \emph{Semi-supervised learning and domain adaptation for NLP}.
\newblock Synthesis Lectures on Human Language Technologies. Morgan \& Claypool
  Publishers, United States.

\bibitem[{Taboada et~al.(2011)Taboada, Brooke, Tofiloski, Voll, and
  Stede}]{taboada-etal-2011-lexicon}
Maite Taboada, Julian Brooke, Milan Tofiloski, Kimberly Voll, and Manfred
  Stede. 2011.
\newblock \href {https://doi.org/10.1162/COLI_a_00049} {Lexicon-based methods
  for sentiment analysis}.
\newblock \emph{Computational Linguistics}, 37(2):267--307.

\bibitem[{Tay et~al.(2020)Tay, Dehghani, Bahri, and Metzler}]{tay2020efficient}
Yi~Tay, Mostafa Dehghani, Dara Bahri, and Donald Metzler. 2020.
\newblock \href {https://arxiv.org/abs/2009.06732} {Efficient transformers: {A}
  survey}.
\newblock \emph{CoRR}, abs/2009.06732.

\bibitem[{Tsarfaty et~al.(2020)Tsarfaty, Bareket, Klein, and
  Seker}]{tsarfaty-etal-2020-spmrl}
Reut Tsarfaty, Dan Bareket, Stav Klein, and Amit Seker. 2020.
\newblock \href {https://doi.org/10.18653/v1/2020.acl-main.660} {From {SPMRL}
  to {NMRL}: What did we learn (and unlearn) in a decade of parsing
  morphologically-rich languages ({MRL}s)?}
\newblock In \emph{Proceedings of ACL 2020}, pages 7396--7408.

\bibitem[{Turc et~al.(2021)Turc, Lee, Eisenstein, Chang, and
  Toutanova}]{Turc2021}
Iulia Turc, Kenton Lee, Jacob Eisenstein, Ming-Wei Chang, and Kristina
  Toutanova. 2021.
\newblock \href {http://arxiv.org/abs/2106.16171} {{Revisiting the Primacy of
  English in Zero-shot Cross-lingual Transfer}}.
\newblock \emph{arXiv preprint arXiv:2106.16171}.

\bibitem[{Turian et~al.(2010)Turian, Ratinov, and
  Bengio}]{turian-etal-2010-word}
Joseph Turian, Lev-Arie Ratinov, and Yoshua Bengio. 2010.
\newblock \href {https://aclanthology.org/P10-1040} {Word representations: A
  simple and general method for semi-supervised learning}.
\newblock In \emph{Proceedings of ACL 2010}, pages 384--394.

\bibitem[{Vania and Lopez(2017)}]{vania-lopez-2017-characters}
Clara Vania and Adam Lopez. 2017.
\newblock \href {https://aclanthology.org/P17-1184} {From characters to words
  to in between: Do we capture morphology?}
\newblock In \emph{Proceedings of ACL 2017}, pages 2016--2027.

\bibitem[{Vanmassenhove et~al.(2021)Vanmassenhove, Shterionov, and
  Gwilliam}]{vanmassenhove-etal-2021-machine}
Eva Vanmassenhove, Dimitar Shterionov, and Matthew Gwilliam. 2021.
\newblock \href {https://aclanthology.org/2021.eacl-main.188} {Machine
  translationese: Effects of algorithmic bias on linguistic complexity in
  machine translation}.
\newblock In \emph{Proceedings of the EACL 2021}, pages 2203--2213.

\bibitem[{Vig et~al.(2020)Vig, Gehrmann, Belinkov, Qian, Nevo, Singer, and
  Shieber}]{vig2020investigating}
Jesse Vig, Sebastian Gehrmann, Yonatan Belinkov, Sharon Qian, Daniel Nevo,
  Yaron Singer, and Stuart~M Shieber. 2020.
\newblock \href
  {https://proceedings.neurips.cc/paper/2020/hash/92650b2e92217715fe312e6fa7b90d82-Abstract.html}
  {Investigating gender bias in language models using causal mediation
  analysis}.
\newblock In \emph{Proceedings of NeurIPS 2020}.

\bibitem[{Vilares et~al.(2018)Vilares, Peng, Satapathy, and
  Cambria}]{Vilares:2018sentic}
David Vilares, Haiyun Peng, Ranjan Satapathy, and Erik Cambria. 2018.
\newblock {BabelSenticNet: A} commonsense reasoning framework for multilingual
  sentiment analysis.
\newblock In \emph{2018 IEEE Symposium Series on Computational Intelligence
  (SSCI)}, pages 1292--1298.

\bibitem[{Volkova et~al.(2013)Volkova, Wilson, and
  Yarowsky}]{volkova-etal-2013-exploring}
Svitlana Volkova, Theresa Wilson, and David Yarowsky. 2013.
\newblock \href {https://aclanthology.org/D13-1187} {Exploring demographic
  language variations to improve multilingual sentiment analysis in social
  media}.
\newblock In \emph{Proceedings of EMNLP 2013}, pages 1815--1827.

\bibitem[{Wang et~al.(2020)Wang, K, Mayhew, and Roth}]{Wang20ExtendmBERT}
Zihan Wang, Karthikeyan K, Stephen Mayhew, and Dan Roth. 2020.
\newblock \href {https://www.aclweb.org/anthology/2020.findings-emnlp.240/}
  {Extending multilingual {BERT} to low-resource languages}.
\newblock In \emph{Findings of EMNLP 2020}, pages 2649--2656.

\bibitem[{Weber(2006)}]{weber2006reach}
Ron Weber. 2006.
\newblock Reach and grasp in the debate over the is core: An empty hand?
\newblock \emph{Journal of the Association for Information Systems}, 7(10):28.

\bibitem[{Weiss et~al.(2018)Weiss, Goldberg, and Yahav}]{Weiss2018}
Gail Weiss, Yoav Goldberg, and Eran Yahav. 2018.
\newblock \href {http://arxiv.org/abs/1711.09576} {{Extracting Automata from
  Recurrent Neural Networks Using Queries and Counterexamples}}.
\newblock In \emph{Proceedings of ICML 2018}.

\bibitem[{Wiegreffe and Pinter(2019)}]{wiegreffe-pinter-2019-attention}
Sarah Wiegreffe and Yuval Pinter. 2019.
\newblock \href {https://aclanthology.org/D19-1002} {Attention is not not
  explanation}.
\newblock In \emph{Proceedings of EMNLP-IJCNLP 2019}, pages 11--20.

\bibitem[{Xu et~al.(2021)Xu, Zhou, Gan, Zheng, and
  Li}]{xu-etal-2021-vocabulary}
Jingjing Xu, Hao Zhou, Chun Gan, Zaixiang Zheng, and Lei Li. 2021.
\newblock \href {https://aclanthology.org/2021.acl-long.571} {Vocabulary
  learning via optimal transport for neural machine translation}.
\newblock In \emph{Proceedings of ACL-IJCNLP 2021}, pages 7361--7373.

\bibitem[{Yang and Eisenstein(2017)}]{Yang:2017tacl}
Yi~Yang and Jacob Eisenstein. 2017.
\newblock \href {https://aclanthology.org/Q17-1021} {Overcoming language
  variation in sentiment analysis with social attention}.
\newblock \emph{Transactions of the Association for Computational Linguistics},
  5:295--307.

\bibitem[{Yarowsky(1995)}]{yarowsky-1995-unsupervised}
David Yarowsky. 1995.
\newblock \href {https://aclanthology.org/P95-1026} {Unsupervised word sense
  disambiguation rivaling supervised methods}.
\newblock In \emph{Proceedings of ACL 1995}, pages 189--196.

\bibitem[{Yin et~al.(2021)Yin, Li, Hu, Peng, and Chang}]{yin-etal-2021-broaden}
Da~Yin, Liunian~Harold Li, Ziniu Hu, Nanyun Peng, and Kai-Wei Chang. 2021.
\newblock \href {https://arxiv.org/pdf/2109.06860.pdf} {Broaden the vision:
  {G}eo-diverse visual commonsense reasoning}.
\newblock In \emph{Proceedings of EMNLP 2021}, pages 2115--2129.

\bibitem[{Zaheer et~al.(2020)Zaheer, Guruganesh, Dubey, Ainslie, Alberti,
  Ontanon, Pham, Ravula, Wang, Yang et~al.}]{zaheer2020big}
Manzil Zaheer, Guru Guruganesh, Kumar~Avinava Dubey, Joshua Ainslie, Chris
  Alberti, Santiago Ontanon, Philip Pham, Anirudh Ravula, Qifan Wang, Li~Yang,
  et~al. 2020.
\newblock \href {https://arxiv.org/pdf/2007.14062.pdf} {Big bird: Transformers
  for longer sequences.}
\newblock In \emph{Proceedings of NeurIPS 2020}.

\bibitem[{Zhao et~al.(2020)Zhao, Mukherjee, Hosseini, Chang, and
  Hassan~Awadallah}]{zhao-etal-2020-gender}
Jieyu Zhao, Subhabrata Mukherjee, Saghar Hosseini, Kai-Wei Chang, and Ahmed
  Hassan~Awadallah. 2020.
\newblock \href {https://aclanthology.org/2020.acl-main.260} {Gender bias in
  multilingual embeddings and cross-lingual transfer}.
\newblock In \emph{Proceedings of ACL 2020}, pages 2896--2907.

\bibitem[{Zhao et~al.(2021)Zhao, Zhu, Shareghi, Vuli{\'c}, Reichart, Korhonen,
  and Sch{\"u}tze}]{Zhao:2021acl}
Mengjie Zhao, Yi~Zhu, Ehsan Shareghi, Ivan Vuli{\'c}, Roi Reichart, Anna
  Korhonen, and Hinrich Sch{\"u}tze. 2021.
\newblock \href {https://aclanthology.org/2021.acl-long.447} {A closer look at
  few-shot crosslingual transfer: {T}he choice of shots matters}.
\newblock In \emph{Proceedings of ACL-IJNCLP 2021}, pages 5751--5767.

\bibitem[{Zhu et~al.(2020)Zhu, Ahuja, Juan, Wei, and
  Reddy}]{zhu-etal-2020-question}
Ming Zhu, Aman Ahuja, Da-Cheng Juan, Wei Wei, and Chandan~K. Reddy. 2020.
\newblock \href {https://doi.org/10.18653/v1/2020.findings-emnlp.342} {Question
  answering with long multiple-span answers}.
\newblock In \emph{Findings of the Association for Computational Linguistics:
  EMNLP 2020}, pages 3840--3849, Online. Association for Computational
  Linguistics.

\end{thebibliography}
\bibliographystyle{acl_natbib}

\appendix

\section{Appendix}

\subsection{Annotation guidelines} \label{app:annotation_guidelines}

For multilinguality, we consider papers that evaluate on 3 languages, or 4 languages if they focus on MT (as the standard MT experiment includes two languages). For fairness and bias, we consider papers that improve fairness in a specific setting or analyze the bias of a method, e.g. regarding gender. For efficiency, we consider papers that analyze memory, speed, or computational complexity. For interpretability, we consider papers that interpret or explain a model's predictions.

In every case, we consider papers that make a \emph{practical contribution} to a dimension and provide quantifiable results along the dimension. For multilinguality, fairness and bias, and efficiency, a practical contribution constitutes the use of an evaluation metric that is appropriate for the specific setting. For interpretability, this may include a user study, an analysis of correlation results, or a qualitative analysis of interpretable features.

\subsection{Analysis of remaining areas at ACL 2021} \label{app:acl2021_remaining}

We provide statistics for the remaining areas at ACL 2021 in Table \ref{tab:acl2021_remaining}.

\begin{table*}[]
\centering
\resizebox{\textwidth}{!}{%
\begin{tabular}{l ccccccccc}
\toprule
Area & \# papers & & English & Accuracy / F1 & Multilinguality & Fairness and bias & Efficiency & Interpretability & \textgreater{}1 dimension \\ \midrule
Question Answering & 24 &  & 95.8\% & 41.7\% & 4.2\% & 4.2\% & 8.3\% & 4.2\% & 0.0\% \\
Sentence-level Semantics & 23 &  & 87.0\% & 56.5\% & 8.7\% & 0.0\% & 4.3\% & 17.4\% & 4.3\% \\
Computational Social Science & 18 &  & 77.8\% & 66.7\% & 0.0\% & 22.2\% & 0.0\% & 16.7\% & 0.0\% \\
Language Generation & 18 &  & 83.3\% & 0.0\% & 11.1\% & 5.6\% & 11.1\% & 11.1\% & 5.6\% \\
Sentiment Analysis & 18 &  & 100.0\% & 72.2\% & 0.0\% & 0.0\% & 11.1\% & 11.1\% & 0.0\% \\
Summarization & 12 &  & 91.7\% & 0.0\% & 0.0\% & 8.3\% & 0.0\% & 8.3\% & 0.0\% \\
Semantics: Lexical Semantics & 12 &  & 58.3\% & 41.7\% & 25.0\% & 0.0\% & 16.7\% & 0.0\% & 8.3\% \\
Information Retrieval & 12 &  & 91.7\% & 8.3\% & 0.0\% & 0.0\% & 0.0\% & 0.0\% & 8.3\% \\
Language Grounding to Vision & 11 &  & 100.0\% & 18.2\% & 0.0\% & 0.0\% & 9.1\% & 27.3\% & 0.0\% \\
Syntax & 10 &  & 40.0\% & 20.0\% & 30.0\% & 0.0\% & 20.0\% & 10.0\% & 20.0\% \\
Best Paper Session & 8 &  & 50.0\% & 50.0\% & 12.5\% & 0.0\% & 25.0\% & 25.0\% & 12.5\% \\
Speech and Multimodality & 6 &  & 66.7\% & 33.3\% & 16.7\% & 0.0\% & 0.0\% & 0.0\% & 0.0\% \\
Phonology and Morphology & 6 &  & 33.3\% & 33.3\% & 33.3\% & 0.0\% & 0.0\% & 16.7\% & 16.7\% \\
Linguistic Theories & 6 &  & 100.0\% & 16.7\% & 0.0\% & 0.0\% & 16.7\% & 33.3\% & 0.0\% \\
Theme & 5 &  & 20.0\% & 40.0\% & 20.0\% & 20.0\% & 20.0\% & 20.0\% & 20.0\% \\
\bottomrule
\end{tabular}%
}
\caption{The number of papers in the remaining areas as well as the fractions that only evaluate on English, only use accuracy / F1, make contributions along one of four dimensions, and make contributions along more than a single dimension (from left to right).}
\label{tab:acl2021_remaining}
\end{table*}

\subsection{Analysis of Efficiency area at EMNLP 2021} \label{app:efficiency_emnlp2021}

We annotated the 20 papers presented orally at EMNLP 2021 in the ``Efficient Models in NLP'' area. Among the presented papers, 19/20 are monolingual and 17 focus only on English. Among the other two, one focuses on Indonesian and one on Chinese. The last paper focuses on MT with multiple languages. Papers mainly evaluate using accuracy and/or F1 and many papers evaluate on GLUE. There is a single two-dimensional paper according to our criteria (the paper focusing on MT, which makes contributions on multilinguality and efficiency) while two other papers can be considered two-dimensional but cover dimensions that we do not annotate, i.e. privacy and robustness respectively. This analysis corroborates our findings that research papers depart from \point in such dedicated conference areas/tracks, but largely only across a single dimension.

\end{document}